\documentclass[letterpaper]{article} 
\usepackage{aaai25}  
\usepackage{times}  
\usepackage{helvet}  
\usepackage{courier}  
\usepackage[hyphens]{url}  
\usepackage{graphicx} 
\usepackage{natbib}  
\usepackage{caption} 
\usepackage{algorithm}
\usepackage{algorithmic}

\usepackage{newfloat}
\usepackage{listings}
\DeclareCaptionStyle{ruled}{labelfont=normalfont,labelsep=colon,strut=off} 
\floatstyle{ruled}
\newfloat{listing}{tb}{lst}{}
\floatname{listing}{Listing}
\usepackage{graphicx}
\usepackage{booktabs}
\usepackage{multirow}
\usepackage{comment}
\usepackage{algorithm}
\usepackage{algorithmic}
\usepackage{float}
\usepackage{subfigure}
\usepackage{capt-of}
\usepackage{colortbl}
\usepackage{amsthm}

\usepackage{xcolor}

\usepackage{mylatexstyle}

\usepackage{amsmath,amsfonts,bm}

\def\eqref#1{equation~\ref{#1}}

\def\1{\bm{1}}

\def\vb{{\bm{b}}}

\DeclareMathAlphabet{\mathsfit}{\encodingdefault}{\sfdefault}{m}{sl}
\SetMathAlphabet{\mathsfit}{bold}{\encodingdefault}{\sfdefault}{bx}{n}

\usepackage{dsfont}
\usepackage{amsthm}

\usepackage{dsfont}

\title{Temporal-Aware Evaluation and Learning for Temporal Graph Neural Networks}
\author {
    Junwei Su\textsuperscript{\rm $\dagger$,*},
    Shan Wu\textsuperscript{\rm $\ddagger$,*}
}
\affiliations {
    \textsuperscript{\rm $\ddagger$}School of Resources and Environmental Engineering, Hefei University of Technology\\
    \textsuperscript{\rm $\dagger$}School of Computing and Data Science, University of Hong Kong\\
    jwsu@cs.hku.hk, wus@hfut.edu.cn 
}

\usepackage{bibentry}

\begin{document}

\maketitle

\begin{abstract}
Temporal Graph Neural Networks (TGNNs) are a family of graph neural networks designed to model and learn dynamic information from temporal graphs. Given their substantial empirical success, there is an escalating interest in TGNNs within the research community. However, the majority of these efforts have been channelled towards algorithm and system design, with the evaluation metrics receiving comparatively less attention. Effective evaluation metrics are crucial for providing detailed performance insights, particularly in the temporal domain. This paper investigates the commonly used evaluation metrics for TGNNs and illustrates the failure mechanisms of these metrics in capturing essential temporal structures in the predictive behaviour of TGNNs. We provide a mathematical formulation of existing performance metrics and utilize an instance-based study to underscore their inadequacies in identifying volatility clustering (the occurrence of emerging errors within a brief interval). This phenomenon has profound implications for both algorithm and system design in the temporal domain. To address this deficiency, we introduce a new volatility-aware evaluation metric (termed volatility cluster statistics), designed for a more refined analysis of model temporal performance. Additionally, we demonstrate how this metric can serve as a temporal-volatility-aware training objective to alleviate the clustering of temporal errors. Through comprehensive experiments on various TGNN models, we validate our analysis and the proposed approach. The empirical results offer revealing insights: 1) existing TGNNs are prone to making errors with volatility clustering, and 2) TGNNs with different mechanisms to capture temporal information exhibit distinct volatility clustering patterns. Moreover, our empirical findings demonstrate that our proposed training objective effectively reduces volatility clusters in error. \let\thefootnote\relax\footnotetext{* corresponding author}
\end{abstract}

\section{Introduction}

Many real-world problems and systems are naturally modeled as \emph{temporal graphs} (also referred to as \emph{dynamic graphs}), characterized by continuously changing relationships, nodes, and attributes. To address this temporal dynamic nature, Temporal Graph Neural Networks (TGNNs), the temporal counterparts to GNNs, have emerged as promising deep learning models capable of modelling time-varying graph structures~\citep{kazemi2020representation,skarding2021foundations,zhang2023tiger,xu2020tgat}. Unlike their static counterparts, TGNNs excel at capturing temporal dependencies and learning temporal representations within the context of temporal graphs. Consequently, they are widely employed in applications such as traffic prediction~\citep{zhao2019t,guo2019attention,zhang2020spatio}, financial analysis~\citep{wang2021temporal,su2024mtrgleffective},  social network~\citep{zhang2021cope}, recommender systems~\citep{kumar2019jodie}, and climate modeling~\citep{khodayar2018spatio}.

Given their substantial empirical success, there is growing interest in TGNNs within the research community. However, most efforts have been concentrated on algorithm and system design, with various classes of TGNNs emerging based on their mechanisms for capturing temporal information (e.g., RNN-based, memory-based, and attention-based; see related work for more details). Conversely, the evaluation of TGNNs has received comparatively less attention. There are only a few benchmark studies on TGNNs that predominantly investigate how various combinations of learning settings and datasets impact the performance of TGNN models. \emph{Notably, these benchmarks typically utilize common instance-based evaluation metrics like Average Precision (AP) and Area Under the ROC Curve (AU-ROC), where each test sample is considered identically and independently}. An intriguing finding from these benchmark studies is that almost all existing TGNNs demonstrate remarkable (and similar) performance when evaluated against these instance-based metrics. \emph{This uniformity in performance poses a significant challenge in model selection for practical applications, as distinguishing between models based on these metrics alone becomes difficult.} Therefore, there is an urgent need to develop more nuanced evaluation metrics that can better capture the unique capabilities and efficiencies of different TGNN architectures.

In addition to model selection, this paper argues that instance-based evaluation metrics are insufficient and ineffective at capturing the temporal structure of the predictive behavior of TGNNs. Data samples in temporal graphs could exhibit \emph{temporal correlation}, impacting the predictions made by TGNNs and introducing patterns such as \emph{volatility clusters}—periods where large fluctuations are grouped together. This aspect is crucial for the functionality of temporal algorithms and systems in TGNNs. For example, in financial trading algorithms or risk management systems, accurately measuring and predicting volatility clusters can be crucial for effective strategy deployment and risk assessment. Similarly, in fault-tolerant systems, understanding volatility clusters can aid in preemptively identifying periods of potential system stress or failure, thereby enabling proactive maintenance or system adjustments to prevent downtime. Adequate performance evaluation ensures that these systems are not only accurate but also robust and responsive under varying temporal dynamics. This, in turn, aids in optimizing operational efficiency, improving decision-making processes, and ensuring reliability in critical applications where timing and the evolution of data play a vital role. Therefore, developing and refining evaluation metrics that can effectively measure the performance of TGNNs is essential for advancing these technologies and their applications.

{\bf Contribution.} This paper aims to spotlight an under-explored aspect of TGNNs—the evaluation metrics. We examine and highlight the inadequacies of current evaluation metrics in capturing the temporal structures of TGNNs and propose a novel performance metric tailored to detect nuanced temporal information such as volatility clusters. The key contributions and findings of this paper are summarized and highlighted as follows

\begin{enumerate}
\item We present a mathematical formulation of existing evaluation metrics alongside a formal definition aimed at measuring the expressiveness of these metrics. This foundational framework is crucial for analyzing evaluation metrics comprehensively and formalizing the limitations inherent in current TGNN evaluation approaches. Utilizing this framework, we formally prove that instance-based evaluation metrics such as AP and AU-ROC resemble a simple counting process and fail to capture temporal structures (e.g., volatility clusters) in the predictions of TGNNs (Theorem~\ref{theorem:failure}).

\item Building on the insights from our analysis, we propose a novel evaluation metric, named \textbf{v}olatility-\textbf{c}luster \textbf{s}tatistics (VCS). Inspired by Hopkins statistics~\citep{hopkins1954new}, VCS serves as a complementary evaluation metric designed to detect and evaluate volatility clusters in the prediction errors of TGNNs. VCS offers crucial insights into the temporal structure of the prediction errors (error pattern) of TGNNs and helps differentiate the performance of various TGNN models.

\item Beyond its use in evaluation, we demonstrate that the concept of VCS can also function effectively as a regularization technique to mitigate volatility clusters in errors with appropriate modifications. We introduce a method termed \textbf{v}olatility-\textbf{c}luster-\textbf{a}ware (VCA) learning, which is a smooth and differentiable extension of VCS. VCA helps mitigate volatility clusters in the prediction errors of TGNNs. This capability is particularly valuable in the design of systems and algorithms for critical areas such as fault-tolerant systems.

\item We validate our findings and the effectiveness of our metrics through extensive empirical studies consisting of five datasets and six SOTA methods. Our empirical results reveal several key insights: 1) existing TGNNs tend to produce volatility cluster in errors, particularly in RNN-based and memory-based models; 2) different types of TGNNs manifest varying error patterns—for instance, memory-based TGNNs generally exhibit clustered errors towards the end of the testing period, whereas RNN-based TGNNs tend to show them at the beginning. These observations indicate fundamental differences in how these models process temporal information and provide directions for model-specific improvements; 3) our proposed VCA learning objective serves as an effective regularization tool, making existing TGNNs less susceptible to volatility clustering in errors.
\end{enumerate}

\begin{figure*}[!t]
\vspace{-6mm}
\centering
\subfigure[TGNN Learning Pipeline]{
\includegraphics[width=.4 \textwidth]{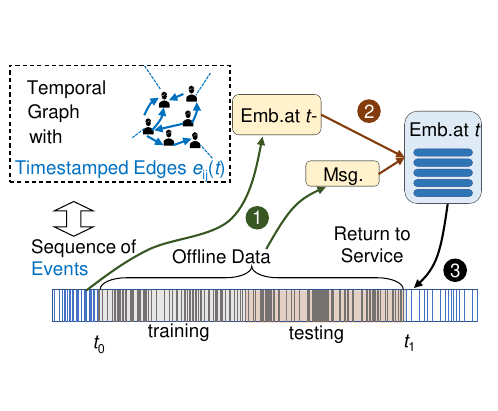}
\label{fig:tgnn_pipe}
}
\subfigure[TGNN Training and Computation]{
\includegraphics[width=.45\textwidth]{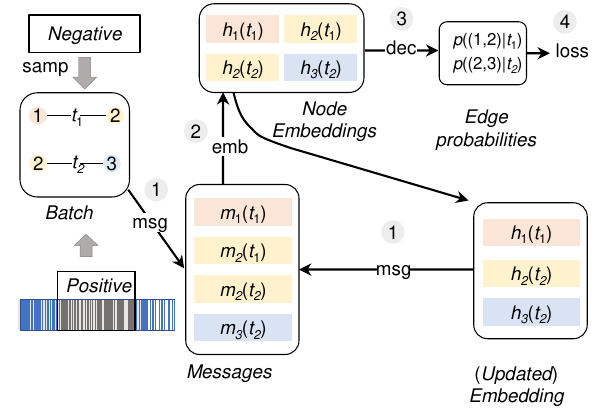}
\label{fig:comp}
}
\vspace{-2mm}
\caption{The Learning Procedure of TGNNs. Fig.~\ref{fig:tgnn_pipe} depicts the learning procedure of TGNN. Data/events are split based on chronological order into training and testing/validation. During the training, data/events are further divided into temporal batches. The incoming batch serves as training samples for updating the model and embedding for the subsequent batch. Fig.~\ref{fig:comp} visualizes the training procedure and computation of TGNNs. Incoming events are served as positive samples and negative events are sampled from the rest of the graphs.}
\label{fig:tgnn}
\end{figure*}

\section{Related Works}\label{sec:related_work}

\paragraph{Temporal Graph Neural Network.}
Temporal graph representation learning has garnered substantial attention in recent years, driven by the imperative to model and analyze evolving relationships and temporal dependencies within temporal graphs (we refer the reader to \cite{skarding2021foundations,kazemi2020representation} for more comprehensive surveys). TGNNs, as temporal counterparts to GNNs, have emerged as promising neural models for temporal graph representation learning\citep{sankar2020dysat, dgb_neurips_D&B_2022, xu2020tgat, su2024pres,wang2021apan,kumar2019jodie,trivedi2019dyrep,zhang2023tiger,pareja2020evolvegcn,trivedi2017know,xu2020inductive,luo2022neighborhood} and have shown SOTA performance in many temporal-related tasks. Roughly speaking, existing TGNNs can be categorized into three types based on the mechanism used for capturing temporal information: RNN-based~\citep{trivedi2019dyrep}, attention-based~\citep{wang2021tcl}, and memory-based TGNNs~\citep{rossi2021tgn}. Due to its potential and practical significance, there has been a recent surge in both theoretical exploration~\citep{souza2022provably} and architectural innovation~\citep{rossi2021tgn,wang2021apan,kumar2019jodie,trivedi2019dyrep,zhang2023tiger} related to TGNNs. In addition, there are works dedicated to optimizing both the inference and training efficiency of TGNNs, employing techniques such as incremental learning~\citep{su2023towards,su2024limitationexperiencereplaygnns}, computation duplication~\citep{wang2023tgopt}, CPU-GPU communication optimization~\citep{zhou2022tgl}, staleness~\citep{sheng2024mspipe}, and caching~\citep{wang2021apan}. Despite all these efforts, the evaluation metrics of TGNNs remain underexplored. In this paper, we address this gap and focus on studying the evaluation metrics of TGNNs.

\paragraph{Evaluation of TGNNs.}
Evaluation is core to machine learning research~\citep{zhang2021understanding}. Because of this, evaluation and benchmarking have been extensively studied in static graph representation learning~\citep{dwivedi2023benchmarking,errica2019fair,hu2020open,lv2021we}. Due to the dynamic nature of temporal graphs, properly evaluating temporal link prediction problems has been challenging and complicated with different issues as documented in~\citep{junuthula2018leveraging,haghani2019systemic,junuthula2016evaluating, dgb_neurips_D&B_2022,huang2024temporal,yu2023towards}. In particular,~\citep{dgb_neurips_D&B_2022,huang2024temporal,yu2023towards} are recent benchmark studies focusing on TGNN evaluation on temporal link prediction. Their studies have revealed that learning settings, such as transductive vs. inductive and negative sampling strategies, play a critical role in properly evaluating TGNNs. In addition, these benchmarks reveal that almost all existing TGNN exhibit remarkable (and similar) performance with respect to the commonly used instance-based evaluation metric, rendering model selection challenging in practice. This has inspired and motivated the central research of this paper.

\section{Preliminary and Background}
In this section, we provide a concise introduction to TGNNs. Due to space limitations, a more detailed description is available in the supplementary material for completeness. We use lowercase letters to denote scalars and graph-related objects, and lower and uppercase boldface letters to denote vectors and matrices, respectively.

\paragraph{Event-based Representation of Temporal Graphs.}
In this paper, we adopt the event-based representation of temporal graphs, as described in previous works~\citep{skarding2021foundations,zhang2023tiger}. A temporal graph $\mathcal{G}$ in this representation consists of a node set $\mathcal{V} = \{1,...,N\}$ and an event set $\mathcal{E} = \{e_{ij}(t)\}$, where $i,j \in \mathcal{V}$. The node set $\mathcal{V}$ represents the entities in the graphs. The event set $\mathcal{E}$ represents a stream of events, with each edge $e_{ij}(t)$ corresponding to an interaction event between node $i$ and node $j$ at timestamp $t \geq 0$. Node features and edge features for $v_i$ and $e_{ij}$ are denoted by $\fb_{i}(t)$ and $\fb_{ij}(t)$, respectively. In the case of non-attributed graphs, we assume $\fb_{i}(t)=\mathbf{0}$ and $\fb_{ij}(t)=\mathbf{0}$, representing zero vectors.

\paragraph{Temporal Graph Neural Networks (TGNNs).}
 TGNNs, extended from the standard GNN to the temporal graph, can be viewed as an embedding function (encoder) for finding the temporal representation of vertices in temporal graphs~\citep{su2024pres,rossi2021tgn}. The learned embedding can then be used as input for different downstream tasks. A canonical formulation of the TGNN encoder is to extend the message-passing scheme from GNNs to include time information. The formulation of TGNNs for learning the representation of vertex $i$ is given by:
\begin{equation*}\label{eq:tgnn}
    \begin{split}
       \hb_i(t) &= \mathrm{emb}(\{\mb_{ij}, j \in \mathcal{N}_i(t)\}),\\
        \mb_{ij}(t) &= \mathrm{msg}(\hb_i(t^-),\hb_j(t^-),\fb_{ij}(t), \fb_{i}(t), \fb_{j}(t), \Delta t ),
    \end{split}
\end{equation*}
where  $\hb_i(t^-)$ and $\hb_j(t^-)$ are the embedding of nodes $i$ and $j$ before time $t$ (i.e., at the time of the previous interaction involving node $i$ or $j$), $\mb_{ij}(t)$ is the message from vertex $j$ to $i$ at time $t$ generated from the event $e_{ij}(t)$, $\mathcal{N}_i(t)$ is the temporal neighbours of nodes $i$ up to time $t$, $h_i(t)$ is temporal embedding/representation of nodes $i$ at time $t$, and $\mathrm{msg(.)}$ (e.g., MLP), and $\mathrm{emb(.)}$ (e.g., GCN) are learnable functions. After obtaining the embeddings $h_i(t)$ and $h_j(t)$ in the prescribed manner, an extra simple MLP layer (or decoder in other forms) can be used for the down-stream tasks.

\paragraph{TGNNs Training and Evaluation}
TGNNs are frequently trained in a self-supervised manner using link prediction tasks~\citep{dgb_neurips_D&B_2022,huang2024temporal}, which are commonly conceptualized as a binary classification problem aimed at predicting whether a link will form between two nodes. Consequently, the performance of TGNNs is often evaluated with respect to their success in link prediction tasks. Therefore, in this paper, we concentrate our discussion on link prediction, though the analysis and arguments can be naturally extended to other downstream tasks such as node classification. More formally, we can assign labels for events $e_{ij}(t)$, such that:
\begin{align*}
    y_{ij}(t) = 
\begin{cases} 
1 & \text{if } e_{ij}(t) \in \mathcal{E}, \\
0 & \text{otherwise}.
\end{cases}
\end{align*}

For simplicity, we omit the specific node pair $i,j$ when referring to the event $e_{ij}(t)$ and index the event by its order of appearance in the corresponding set. Let $\mathcal{E}_{\mathrm{test}} = \{e_{k}(t_k)\}_{k=1,...,M},$
be a chronologically ordered sequence of $M$ test samples from the test period, $T_{\mathrm{test}} = [t_1,t_2], i.e., t_1 \leq t_k \leq t_{k+1} \leq t_2.$ 
Let $\Yb = \{y_1,...,y_m\},$ be the ground-truth labels of the given samples, and let $\hat{\Yb} = \{\hat{y}_1,...,\hat{y}_m\},$ be the predicted labels of the given samples by the TGNN. Then, we can define the performance evaluation metric  as a function $\mu(.)$ of the form:
$$\mu: \Yb \times \hat{\Yb} \times \cE \mapsto \RR^+.$$
In other words, $\mu$ takes in the prediction and the ground truth and maps them to a positive real value.

\begin{figure*}[!t]
    \vspace{-5mm}
    \centering
   \subfigure[Random Error]{
\includegraphics[width=.25\textwidth]{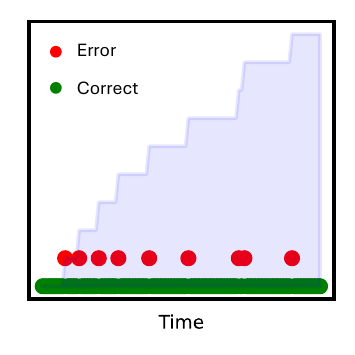}
\label{fig:random_err}
}
   \subfigure[Cluster Error]{
\includegraphics[width=.25\textwidth]{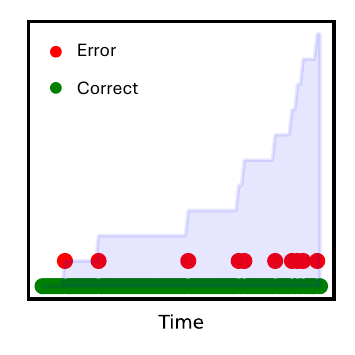}
\label{fig:cluster_err}
}
   \subfigure[Regular Error]{
\includegraphics[width=.25\textwidth]{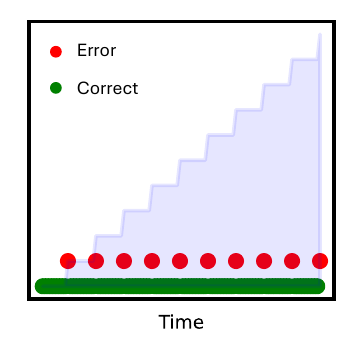}
\label{fig:regular_err}
}
\vspace{-2mm}
  \caption{An illustration of different error patterns. Fig.~\ref{fig:random_err} is the pattern for random error pattern where wrong predictions are randomly distributed across the time interval. Fig.~\ref{fig:cluster_err} is the pattern for volatility cluster error where wrong predictions are clustered at a small time interval (the end of the temporal horizon in the example).  Fig.~\ref{fig:cluster_err} is the pattern for regular error where wrong predictions are evenly spaced. The shaded area in the plots indicates the accumulated count of errors.} 
  \label{fig:example}
\end{figure*}

\subsection{Limitation of Current Evaluation Metrics} 
To explore the limitation of the evaluation metric, we first define a measure of its capability. In this paper, we propose extending the idea of the expressive power of GNNs to characterize the ability of an evaluation metric by its expressiveness—the capacity to differentiate between different predictions. More formally, we introduce the following definition.
\begin{definition}[Expressiveness of Evaluation Metric]\label{def:expressiveness}
For two distinct predictions $\hat{\Yb}_1$ and $\hat{\Yb}_2$, we say an evaluation metric $\mu$ can differentiate $\hat{\Yb}_1$ and $\hat{\Yb}_2$ if
$
\mu(\Yb,\hat{\Yb}_1,\mathcal{E}) \neq \mu(\Yb,\hat{\Yb}_2,\mathcal{E}).
$
\end{definition}
As noted, the most commonly used evaluation metrics for TGNNs are instance-based, such as AP and AU-ROC, where each test sample is considered identically and independently. More formally, this family of evaluation metrics is defined as follows:
\begin{definition}[Instance-based Evaluation]\label{def:instance_evaluation}
    For a given evaluation $\mu(\Yb, \hat{\Yb}, \cE)$, we say $\mu(.)$ is an instance-based evaluation metric if it can be expressed as, 
    $$\mu(\Yb, \hat{\Yb}, \cE) = g\left ( \left \{f(y_i,\hat{y}_i)|y_i, \hat{y}_i \in \Yb, \hat{\Yb} \right \} \right),$$
    where $g$ is some set function and $f: \Yb \times \hat{\Yb} \mapsto \RR^+$.
\end{definition}

The following result shows the limitation of instance-based evaluation metrics:
\begin{theorem}[Failure of Instance-Based Evaluation]\label{theorem:failure}
    Let $\hat{\Yb}_1$ and $\hat{\Yb}_2$ be two distinct predictions for the set $\cE$ with ground-truth $\Yb$, and $\mu(.)$ is an instance-based evaluation metric. Then, we have that,
    $$\mu(\hat{\Yb}_1, \Yb, \cE) = \mu(\hat{\Yb}_2, \Yb, \cE),$$
    so long as,
    $$\mathrm{H}(\Yb,\hat{\Yb}_1) = \mathrm{H}(\Yb,\hat{\Yb}_2),$$
    where $\mathrm{H}(\Yb,\hat{\Yb}) = \sum_{k=1}^{|\cE|} \mathds{1}[y_k \neq \hat{y}_k]$.
\end{theorem}
\let\thefootnote\relax\footnotetext{Extra technical details such as proof, pseudo-code and experimental setting can be found in extended arxiv version~\citep{su2024temporalawareevaluationlearningtemporal}.}
Theorem~\ref{theorem:failure} demonstrates that instance-based evaluation metrics cannot differentiate predictions if the number of disagreements with the ground truth is the same. Essentially, such metrics reduce all diverse information (e.g., temporal information) of predictions to a mere disagreement count. This severely limits the expressiveness of these metrics, making them inadequate for capturing insightful information about predictions within the temporal process.

\paragraph{Visualization Example.} To further illustrate this, consider the examples in Fig.~\ref{fig:example}, which have identical numbers of errors and correct predictions. It is evident that the instance-based evaluation metric fails to differentiate these examples, as they exhibit the same predictive performance (i.e., the same amount of disagreement/errors). However, the patterns of errors in these examples are markedly different. Such variances in error distribution provide crucial insights into both the TGNN models and the systems they represent. For example, as previously discussed, the presence of a volatility cluster in errors is critical information for model selection in real-time fault-tolerant systems, where functionality is ensured if errors are evenly distributed. Thus, the inability to detect such error patterns can lead to catastrophic failures in many real-world algorithm and system designs. To address this issue, in the subsequent section, we introduce a novel evaluation metric and learning objective designed to detect and mitigate this type of volatility cluster in errors.

\section{Methodology}
Building on the previous discussion regarding the limitations of existing evaluation metrics, this section introduces a novel temporal-aware evaluation metric derived from the concept of Hopkins statistics~\citep{banerjee2004validating}. Specifically, we focus on detecting volatility clusters within predictions, which have significant implications for algorithms and systems, as discussed earlier. Additionally, based on this proposed evaluation metric, we introduce a novel temporal-aware learning objective for TGNNs. 

\subsection{Volatility-Cluster Statistics (VCS)}

Given a test period \(T_{\mathrm{test}}\), let \(\Yb\) and \(\hat{\Yb}\) represent the ground truth and the predictions of the model on the test set, respectively. Let \(\mathcal{E}_{\mathrm{disg}}\) denote the set of disagreement events with cardinality \(K\) and let \(\hat{\mathcal{E}}_{\mathrm{disg}}\) denote $k < K$ samples from \(\mathcal{E}_{\mathrm{disg}}\) . We first compute the sum of distances from the sampled disagreement set to the disagreement as:
\begin{align}
    \mathrm{D}_{\mathrm{disg}} &= \sum_{e \in \hat{\mathcal{E}}_{\mathrm{disg}}}\mathrm{d}(e,\mathcal{E}_{\mathrm{disg}}), \\
    \mathrm{d}(e,\mathcal{E}_{\mathrm{disg}}) &= \min \left\{|t_e-t_{e'}| \bigg| e' \in \mathcal{E}_{\mathrm{disg}}, e' \neq e \right\}.
\end{align}
\(\mathrm{d}(e,\mathcal{E}_{\mathrm{disg}})\) calculates the time difference between event \(e\) and the closest event in the given set. Then, \(\mathrm{D}_{\mathrm{disg}}\) is a sum of such distances for the disagreement set. Next, we generate a set \(\mathcal{E}_{\mathrm{r}}\) of \(k\) events by uniformly randomly sampling from the test period \(T_{\mathrm{test}}\). Similarly, we compute its distance to the disagreement as:
\begin{align}
    \mathrm{D}_{\mathrm{r}} &= \sum_{e \in \mathcal{E}_{\mathrm{r}}}\mathrm{d}(e,\mathcal{E}_{\mathrm{disg}}).
\end{align}
$\mathrm{D}_{\mathrm{r}}$ serves as a reference for the distance to the disagreement if the samples are randomly drawn. Then, we can compute relative statistics between the set \(\mathcal{E}_{\mathrm{disg}}\) and \(\mathcal{E}_{\mathrm{r}}\) as:
\begin{align*}
    \mathcal{T}(\mathcal{E}_{\mathrm{disg}}, \mathcal{E}_{\mathrm{r}}) = \frac{\mathrm{D}_{\mathrm{r}}}{\mathrm{D}_{\mathrm{r}}+\mathrm{D}_{\mathrm{disg}}},
\end{align*}
where \(\mathrm{D}_{\mathrm{r}}\) and \(\mathrm{D}_{\mathrm{disg}}\) are described above.  The formulation shows that \(\mathcal{T}(\mathcal{E}_{\mathrm{disg}}, \mathcal{E}_{\mathrm{r}})\) compares the temporal distance between predictions relative to random sampling. The ratio format confines the value within the range of \(0\) to \(1\). The \(\mathcal{T}(\mathcal{E}_{\mathrm{disg}}, \mathcal{E}_{\mathrm{r}})\) statistic provides insights into the distribution of data points. If \(\mathcal{T}(\mathcal{E}_{\mathrm{disg}}, \mathcal{E}_{\mathrm{test}})\) is close to \(1\), it indicates that the data points are clustered, with the sum of distances from randomly generated points to their nearest neighbors being significantly larger than that from the sampled data points. Conversely, if \(\mathcal{T}(\mathcal{E}_{\mathrm{disg}}, \mathcal{E}_{\mathrm{test}})\) is close to \(0\), it could suggest that the data points are regularly-spaced, resulting in smaller distances for randomly generated points compared to those from sampled data points. When \(\mathcal{T}(\mathcal{E}_{\mathrm{disg}}, \mathcal{E}_{\mathrm{test}})\) approximates \(0.5\), it indicates a random distribution with no significant clustering or regular pattern, as both randomly generated points and sampled data points exhibit similar nearest neighbour distances.

To enhance interoperability and robustness against variance from sampling, we repeat the sampling steps multiple times and adjust based on the random sampling. The final VCS is computed as follows:
\begin{align*}
    \mathrm{VCS} = | 1/2 -
     \mathcal{T}(\mathcal{E}_{\mathrm{disg}}, \mathcal{E}_{\mathrm{r}}, \tau)|,\\ 
     = \left | \frac{1}{2} - \frac{1}{\tau} \sum_{i  = 1}^{\tau} \frac{\mathrm{D}_{\mathrm{r}}^{(i)}}{\mathrm{D}_{\mathrm{r}}^{(i)}+\mathrm{D}_{\mathrm{disg}}^{(i)}} \right |.
\end{align*}
where \(\tau\) is the number of repeated samples. Our empirical study suggests that \(\tau = 5\) provides a stable estimate in most cases.

\subsection{Volatility-Cluster-Aware (VCA) Learning}

In the previous section, we introduced a new statistical measure for detecting volatility clusters in the temporal dimension. We discussed how the error pattern of the system can have significant implications in real-time systems, especially concerning fault-tolerant aspects of development. Typically, real-time systems prefer more uniform error distributions. Thus, an important question arises: can we use the proposed measure to help TGNNs learn a model (weight) from the hypothesis space that exhibits a more uniform error pattern?

A straightforward idea is to incorporate $\mathcal{T}(\mathcal{E}_{\mathrm{disg}}, \mathcal{E}_{\mathrm{test}},\tau)$ as a regularization term in the learning objective. However, a technical challenge arises due to the non-differentiability of the distance function $\mathrm{d}(e,\cE)$, which is due to the $\min$ operator. To address this, we propose the following modification with a smooth and differentiable version that mimics the min function:
\begin{align*}
\mathrm{d}_{\mathrm{soft}}(e,\cE) = -\log \left( \sum_{e' \in \cE, e' \neq e} \exp(-\beta |t_e - t_{e'}|) \right)/\beta,
\end{align*}
where $\beta$ is a positive parameter that controls the sharpness of the approximation. As $\beta$ increases, the approximation becomes closer to the actual minimum function. This approach turns the non-differentiable minimum into a differentiable function by summing over exponentially scaled, inverted distances,
\begin{align*}
\mathcal{T}_{\mathrm{soft}}(\mathcal{E}_{\mathrm{disg}}, \mathcal{E}_{\mathrm{r}}) = \frac{\hat{\mathrm{D}}_{\mathrm{r}}}{\hat{\mathrm{D}}_{\mathrm{r}}+\hat{\mathrm{D}}_{\mathrm{disg}}},
\end{align*}
where $\hat{\mathrm{D}}_{\mathrm{r}}$ and $\hat{\mathrm{D}}_{\mathrm{disg}}$ are defined similarly as before with the distance function replaced with $\mathrm{d}_{\mathrm{soft}}(.)$. We can then incorporate this into the learning process and term the modified objective VCA.
\begin{align}\label{eq:temp_train}
\hat{\mathcal{L}}(\hat{\Yb},\Yb) = \mathcal{L}(\hat{\Yb},\Yb) + \gamma \left \| \frac{1}{2} - \mathcal{T}_{\mathrm{soft}}(\mathcal{E}_{\mathrm{disg}}, \mathcal{E}_{\mathrm{r}}) \right \|^2,
\end{align}
where $\mathcal{L}(\hat{\Yb},\Yb)$ is the standard loss function for training TGNNs (e.g., cross-entropy), and $\gamma$ is a hyper-parameter controlling the regularization effect. If the error pattern deviates from a uniform distribution, then VCA will incur a larger value, and consequently, the training objective will reflect a larger loss. Achieving a lower value with this new training objective is expected to improve the uniformity of the error distribution within the model.

\begin{figure*}[!t]
\vspace{-5mm}
    \centering
   \subfigure[Memory-based TGNN]{
\includegraphics[width=.315\textwidth]{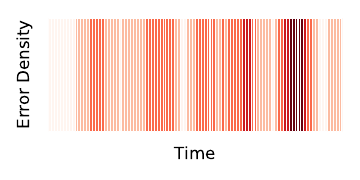}
\label{fig:memory}
}
   \subfigure[RNN-based TGNN]{
\includegraphics[width=.315\textwidth]{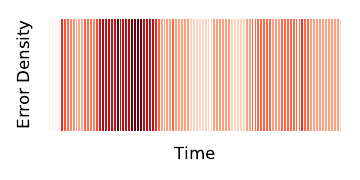}
\label{fig:rnn}
}
   \subfigure[Attention-based TGNN]{
\includegraphics[width=.315\textwidth]{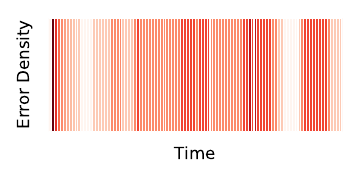}
\label{fig:transformer}
}
\vspace{-3mm}
  \caption{An illustration of the error patterns across different types of TGNNs. The x-axis represents the time during the test period, and the color density indicates the error density (number of errors per time unit). A higher density (redder) indicates more errors. As shown in the figures, memory-based TGNNs exhibit a higher error density toward the end of the testing period, while RNN-based TGNNs display a higher error density at the beginning of the testing period. Attention-based TGNNs, on the other hand, demonstrate a more uniform error distribution.} 
  \label{fig:error_pattern}
\end{figure*}

\begin{table*}[!t]
    \centering
    \scriptsize
    \resizebox{1\textwidth}{!}{
    \begin{tabular}{|l|l|l|l|l|l|l|l|l|l|l|}
    \toprule
Dataset & \multicolumn{2}{c}{Reddit} & \multicolumn{2}{c}{Wikipedia} & \multicolumn{2}{c}{MOOC} & \multicolumn{2}{c}{LastFM} & \multicolumn{2}{c}{GDELT} \\
\hline
Model/Metric & VCS $\downarrow$ & AP(\%) $\uparrow$ & VCS $\downarrow$ & AP(\%) $\uparrow$ & VCS $\downarrow$ & AP(\%) $\uparrow$ & VCS $\downarrow$ & AP(\%) $\uparrow$ & VCS $\downarrow$ & AP(\%) $\uparrow$ \\
\hline
TGN & 0.18±0.02 & 98.5±0.04 & 0.21±0.04 & 96.4±0.03 & 0.25±0.03 & 97.6±0.03 & 0.22±0.04 & 75.4±0.06 & 0.24±0.03 & 95.6±0.05 \\
TGN-VCA & \textbf{0.08±0.01} & 98.2±0.03 & \textbf{0.12±0.02} & 96.3±0.04 & \textbf{0.13±0.03} & 97.3±0.02 & \textbf{0.09±0.03} & 73.3±0.05 & \textbf{0.12±0.02} & 96.8±0.03 \\
\hline
Tiger & 0.23±0.01 & 97.5±0.08 & 0.23±0.03 & 94.8±0.06 & 0.30±0.02 & 95.1±0.04 & 0.23±0.03 & 77.7±0.05 & 0.23±0.03 & 97.5±0.03 \\
Tiger-VCA & \textbf{0.10±0.01} & 98.0±0.06 & \textbf{0.11±0.02} & 94.0±0.06 & \textbf{0.11±0.01} & 95.6±0.03 & \textbf{0.12±0.02} & 78.0±0.04 & \textbf{0.11± 0.01} & 97.0±0.05 \\
\hline
JOIDE & 0.19±0.03 & 96.5±0.05 & 0.25±0.04 & 95.3±0.04 & 0.21±0.03 & 97.5±0.08 & 0.20±0.03 & 72.5±0.06 & 0.27±0.04 & 96.8±0.05 \\
JOIDE-VCA & \textbf{0.09±0.02} & 96.8±0.03 & \textbf{0.11±0.03} & 94.8±0.05 & \textbf{0.11±0.02} & 97.8±0.06 & \textbf{0.10±0.02} & 72.8±0.07 & \textbf{0.13±0.03} & 97.0±0.04 \\
\hline
DyRep & 0.25±0.03 & 96.7±0.06 & 0.22±0.04 & 94.8±0.03 & 0.23±0.03 & 96.8±0.06 & 0.27±0.03 & 69.5±0.05 & 0.24± 0.04 & 97.8±0.03 \\
DyRep-VCA & \textbf{0.11±0.02} & 97.0±0.05 & \textbf{0.10±0.03} & 95.0±0.04 & \textbf{0.12±0.02} & 97.0±0.05 & \textbf{0.12±0.01} & 70.0±0.06 & \textbf{0.14±0.03} & 97.5±0.04 \\
\hline
TCL & 0.12±0.02 & 95.5±0.02 & 0.11±0.02 & 91.6±0.06 & 0.14±0.03 & 93.5±0.07 & 0.14±0.03 & 68.5±0.07 & 0.14±0.03 & 94.6±0.06 \\
TCL-VCA & \textbf{0.09±0.02} & 95.2±0.02 & \textbf{0.06±0.01} & 92.2±0.06 & \textbf{0.10±0.02} & 92.8±0.05 & \textbf{0.09±0.01} & 67.5±0.03 & \textbf{0.10±0.0} & 95.2±0.06 \\
\hline
TGAT & 0.13±0.02 & 95.8±0.03 & 0.10±0.01 & 92.3±0.03 & 0.14±0.03 & 94.3±0.03 & 0.13±0.03 & 70.1±0.05 & 0.12±0.03 & 93.3±0.03 \\
TGAT-VCA & \textbf{0.10±0.02} & 96.0±0.02 & \textbf{0.08±0.01} & 93.0±0.04 & \textbf{0.07±0.02} & 95.0±0.05 & \textbf{0.06±0.01} & 71.3±0.06 & \textbf{0.09±0.02} & 93.0±0.04\\
\hline
$\Delta$VCS & \cellcolor{green}0.09 & & \cellcolor{green}0.09 & & \cellcolor{green}0.1 & & \cellcolor{green}0.1 & &  \cellcolor{green}0.09\\
\bottomrule
\end{tabular}
    }
    \caption{The VCS of TGNNs with and without the VCA learning objective. The experiment follows the standard setting. Models labelled with $.$-VCA are trained using our proposed learning objective as defined in Eq.~\ref{eq:temp_train}, with $\tau = 5$ and $\gamma = 0.1$. $\downarrow$ indicates that smaller values are better, while $\uparrow$ indicates that larger values are better. The bolded entry indicate improvement with VCA.  The last row $\Delta$ shows the average improvement with VCA for each dataset. The results in this table collectively demonstrate that VCS can successfully detect volatility clusters in errors, and VCA is effective in mitigating them.}
    \label{tab:cluster}
\end{table*}

\begin{figure*}[!t]
    \centering
   \subfigure[$\tau$ vs Variance]{
\includegraphics[width=.28\textwidth]{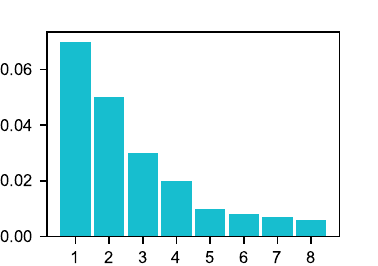}
\label{fig:tau_var}
}
   \subfigure[$\gamma$ vs VCS]{
\includegraphics[width=.28\textwidth]{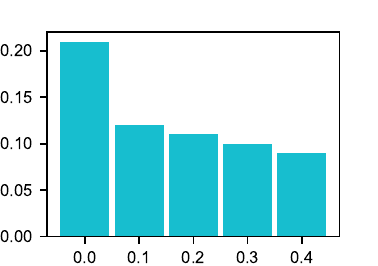}
\label{fig:gamma_vcs}
}
   \subfigure[$\gamma$ vs AP]{
\includegraphics[width=.28\textwidth]{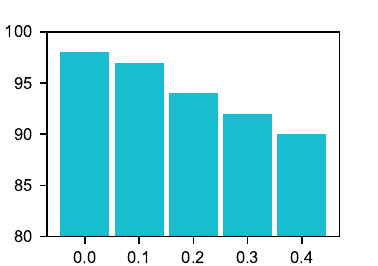}
\label{fig:gamma_ap}
}
  \vspace{-3mm}
  \caption{An illustration of the effects of the hyper-parameters $\tau$ and $\gamma$ on VCS and VCA. Fig.~\ref{fig:gamma_vcs} and ~.\ref{fig:gamma_ap} demonstrate that as $\gamma$ increases, VCS performance improves while AP decreases. Hence, $\gamma$ serves as a control variable that manages the trade-off between VCS and AP. Fig.~\ref{fig:tau_var} shows that increasing $\tau$ reduces the variance in the measure, but the marginal gain diminishes after $\tau = 5$.} 
  \label{fig:error_pattern}
\end{figure*}

\section{Empirical Study}\label{sec:exp}
In this section, we present an empirical study to further illustrate the problem addressed in this paper. The study aims to answer the following key questions:
\begin{enumerate}
\item Do existing TGNNs exhibit volatility clusters in errors?
\item Do existing TGNNs exhibit different error distributions?
\item Is VCS effective in detecting volatility clusters in errors?
\item Can VCA mitigate volatility clusters in errors?
\end{enumerate}

\subsection{Experimental Settings}
\paragraph{Datasets and Baselines.}
We use five public dynamic graph benchmark datasets: Reddit, Wikipedia, MOOC, LastFM, and GDELT~\citep{dgb_neurips_D&B_2022}. We evaluate six state-of-the-art TGNN models, with two models from each of the three categories of TGNNs mentioned: TGN~\citep{rossi2021tgn} \& Tiger~\citep{zhang2023tiger} (memory-based TGNNs), TCL~\citep{wang2021tcl} \& TGAT~\citep{xu2020tgat} (attention-based TGNNs), and JOIDE~\citep{kumar2019jodie} \& DyRep~\citep{trivedi2019dyrep} (RNN-based TGNNs). We adopt the implementation of these baselines from~\cite{zhou2022tgl,dgb_neurips_D&B_2022,huang2024temporal}.

\paragraph{Evaluation Task and Metrics.}
Following the approaches outlined in \citep{dgb_neurips_D&B_2022,huang2024temporal,yu2023towards}, we evaluate models for temporal link prediction, which involves predicting the probability of a link forming between two nodes at a specific time. We use a multi-layer perceptron (MLP) that takes the concatenated representations of two nodes as input and outputs the probability of a link. For evaluation metrics, we focus on AP and the proposed VCS. We train each model with and without VCA to observe the effect of our proposed learning objective. For all experiments, we follow the standard procedure and split datasets chronologically with a ratio of 70\%/15\%/15\% for training, validation, and testing, respectively. Each experiment is conducted with five independent trials, and the average results are reported

\subsection{Experimental Results}\label{subsec:exp_result}

\paragraph{Temporal Error Pattern.}
Our first experiment aims to demonstrate the temporal error patterns of various models and how our proposed metrics can effectively differentiate and reveal insightful information regarding these patterns. Fig.~\ref{fig:error_pattern} illustrates that different types of TGNNs exhibit distinct error pattern behaviours. Specifically, memory-based TGNNs tend to produce volatility clusters in errors toward the end of the test period, RNN-based TGNNs are more prone to errors at the beginning of the test period and attention-based TGNNs exhibit a more uniform distribution in errors. This temporal structure in the prediction errors of memory-based and RNN-based TGNNs is reflected by a larger VCS value in Table~\ref{tab:cluster}
. This confirms that existing TGNNs indeed generate volatility clusters in errors, and different TGNN mechanisms induce varying volatility patterns. Furthermore, this demonstrates that VCS is an effective measure for detecting volatility clusters in errors.

\paragraph{Effectiveness of VCA.}
Our next experiment aims to demonstrate the effectiveness of our proposed learning objective, VCA, as defined in Eq.~\ref{eq:temp_train}, in regulating the behavior of TGNNs. As shown in Table~\ref{tab:cluster}, TGNN models trained with our proposed objective significantly reduce volatility clusters in errors, as evidenced by decreased VCS values. The improvement in attention-based TGNNs (e.g., TCL \& TGAT) is relatively small because these models already exhibit a fairly uniform error distribution. This confirms that VCA is indeed effective in mitigating volatility clusters in errors. Such a property can be particularly beneficial for critical real-time systems where fault tolerance is important, and a more uniformly distributed error is preferred.

\paragraph{Ablation Study.}
The final part of the empirical study focuses on the hyper-parameters of VCS and VCA. The key hyper-parameter in VCS is $\tau$, which represents the number of independent trials conducted to compute the reference distance for random errors. As shown in Fig.~\ref{fig:tau_var}, we found that increasing $\tau$ leads to a smaller variance in value but incurs a higher computational cost. However, we find that $\tau=5$ already provides a sufficiently robust estimation. The main hyper-parameter in VCA is $\gamma$ in Eq.\ref{eq:temp_train}, which controls the regularization effect of the proposed learning objective. Our experiment shows that increasing $\gamma$ results in a more uniform error pattern but worsens predictive performance (smaller AP). Thus, there is a trade-off between achieving this uniform error distribution and maintaining predictive performance. This trade-off does not undermine the effectiveness of our proposed learning objective, as the primary goal is to make the error distribution more uniform. Whether this trade-off is favourable depends on the application scenario. However, as indicated in Table~\ref{tab:cluster}, $\gamma=0.1$ provides a significant improvement in VCS without significantly affecting the model's accuracy.

\section{Discussion}
\paragraph{Conclusion.} In this paper, we investigate the evaluation metrics for TGNNs. Specifically, we have identified the pitfalls and limitations of currently used instance-based measures, such as AP and AU-ROC, in capturing temporal structures in prediction errors, such as volatility clusters. To address this issue, we propose VCS, a metric that effectively captures volatility clusters in errors for TGNNs. Furthermore, we extend this proposed evaluation metric as a regularizer, introducing VCA to mitigate volatility clusters in errors.

\subsection{ Limitation and future works}
In this paper, we focus on volatility clusters in errors. However, other important temporal structures, such as the time of arrival of errors, are not captured by the current metric. We discuss potential approaches for capturing this information in the supplementary material, presenting an interesting avenue for future exploration. Additionally, our study primarily concentrates on the temporal aspect of error distribution. A natural application of TGNNs is in spatio-temporal networks, where vertices represent physical locations, incorporating a spatial dimension. It would be intriguing to explore whether similar concepts can be extended to examine the spatial aspects of TGNNs in spatio-temporal graph networks. This represents another promising area for future research.

\section*{Acknowledgements}
We thank our anonymous reviewers for the valuable feedbacks. This research is supported by the Natural Science Foundation of China (No. 42302326), the Anhui Province Key Research and Development Plan project (No.2022107020029)

\bibliography{reference}

\clearpage
\appendix
\section{Proof for the Failure of Existing Evaluation Metric}
In this appendix, we provide a proof for Theorem~\ref{theorem:failure}. 
\begin{proof}
     Let $\hat{\Yb}_1$ and $\hat{\Yb}_2$ be two distinct predictions for the set $\cE$ with ground-truth $\Yb$ with  $$\mu(\hat{\Yb}_1, \Yb, \cE) = \mu(\hat{\Yb}_2, \Yb, \cE),$$
     where  $$\mathrm{H}(\Yb,\hat{\Yb}) = \sum_{k=1}^{|\cE|} \mathds{1}[y_k \neq \hat{y}_k].$$
     
     Let $\mu(.)$ be an given instance-based evaluation metric. By definition~\ref{def:instance_evaluation}, we can rewrite $\mu(\hat{\Yb}_1, \Yb, \cE), \mu(\hat{\Yb}_2, \Yb, \cE)$ as,
     $$\mu(\hat{\Yb}_1, \Yb, \cE) = g\left ( \left \{f(y_i,\hat{y}_i)|y_i, \hat{y}_i \in \Yb, \hat{\Yb}_1 \right \} \right),$$
          $$\mu(\hat{\Yb}_2, \Yb, \cE) = g\left ( \left \{f(y_i,\hat{y}_i)|y_i, \hat{y}_i \in \Yb, \hat{\Yb}_2 \right \} \right).$$
    As link-prediction problem can be reduced to a binary classification problem, this means that $f(.)$ can be written as,
    $$f(y_i,\hat{y}_i) = c \cdot \mathds{1}[\hat{y}_i \neq y_i],$$
    where $c$ is some constant that weight the wrong prediction. 

    Without loss of generality, we assume $$\mathrm{H}(\Yb,\hat{\Yb}_1) = \mathrm{H}(\Yb,\hat{\Yb}_2) = k,$$
    for some positive integer $k$. Then, we can rewrite $\mu(\hat{\Yb}_1, \Yb, \cE), \mu(\hat{\Yb}_2, \Yb, \cE)$ as,
    \begin{align*}
        \mu(\hat{\Yb}_1, \Yb, \cE) &=  g\left ( \left \{f(y_i,\hat{y}_i)|y_i, \hat{y}_i \in \Yb, \hat{\Yb}_1 \right \} \right) \\
        & = g(k \cdot c ) \\
        & = g\left ( \left \{f(y_i,\hat{y}_i)|y_i, \hat{y}_i \in \Yb, \hat{\Yb}_2 \right \} \right) \\
        & = \mu(\hat{\Yb}_2, \Yb, \cE)
    \end{align*}
\end{proof}
\begin{figure*}[!t]
\centering
\includegraphics[width=1\textwidth]{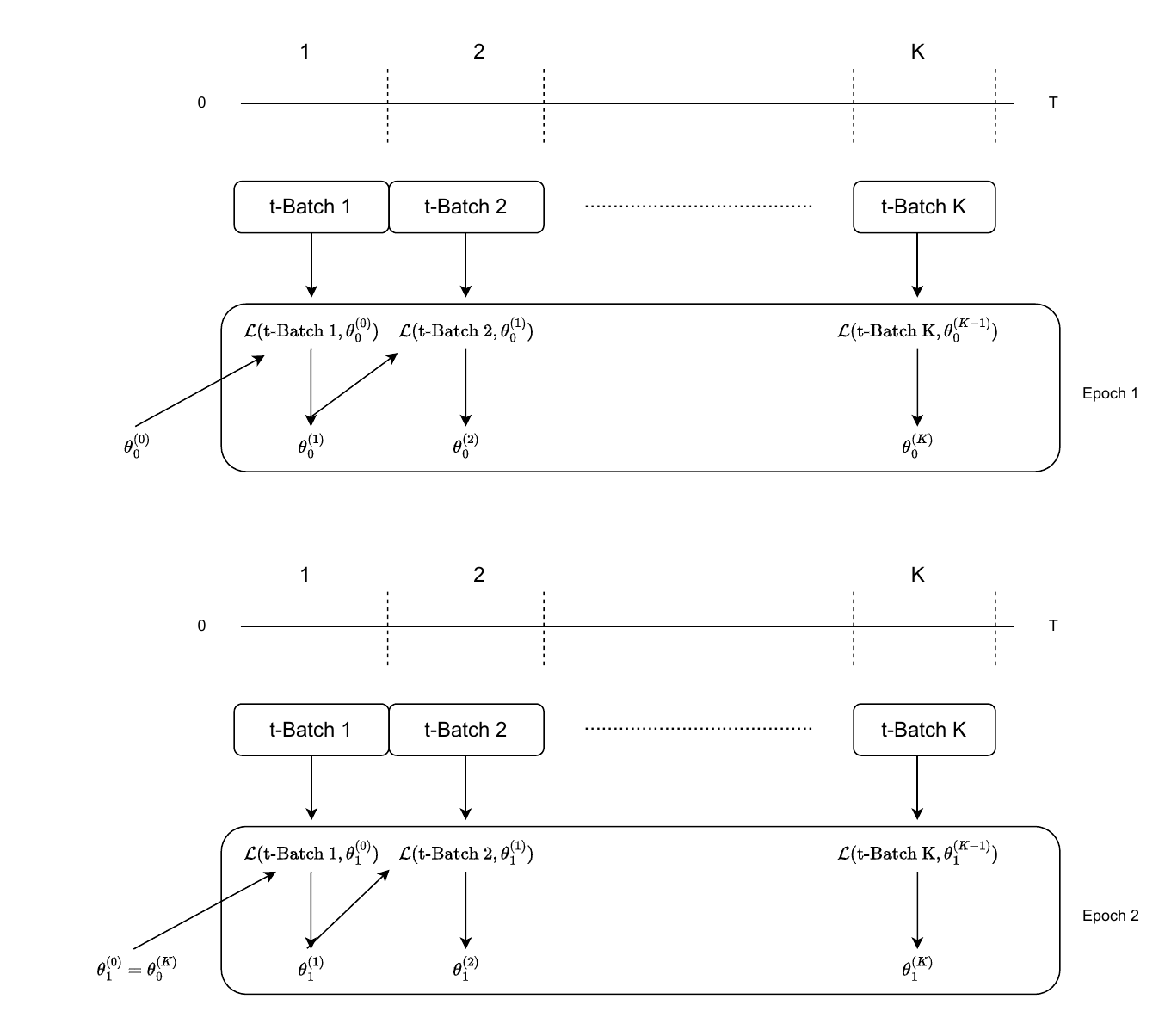}
\caption{Illustration of TGNN Training Procedure. The figure depicts the training flow of TGNN for two epochs.
The incoming batch serves as training samples for updating the model and updating the embedding for the subsequent batch. The model parameter is carried through the second epoch.}
\label{fig:train_dynamic}
\end{figure*}

\section{Algorithm and Further Discussion}\label{appendix:tgnn}
\subsection{Training Procedure of TGNNs}
The training procedure of an TGNN involves capturing temporal dynamics and learning representations of nodes in a dynamic graph. The process typically follows an encoder-decoder framework. In the encoder, the TGNN takes a dynamic graph as input and generates dynamic representations of nodes. This is achieved by using message-passing mechanisms to propagate information through the graph and incorporating temporal neighbors' interactions. The decoder utilizes the node representations generated by the encoder to perform downstream tasks, such as temporal link prediction or node classification. TGNNs are commonly trained using a self-supervised temporal link prediction task, where the decoder predicts the likelihood of an edge between two nodes based on their representations.

The training procedure of an TGNN involves several steps. First, the dataset is divided into training, validation, and test sets using a chronological split. Specifically, given an event set from time interval $[0,T]$, the chronological split partitions the dataset into $[0,T_{\mathrm{train}}]$ (training set), $[T_{\mathrm{train}}, T_{\mathrm{validation}}]$ (validation set), and $[T_{\mathrm{validation}}, T_{\mathrm{test}}]$ (test set). From now on, we focus on the training set and drop the subscript. The training set is then further divided into temporal batches, where each batch consists of consecutive events in the dynamic graph. Additionally, negative events are sampled from the rest of the graph to provide the negative signal. During training, TGNNs often adopt a lag-one procedure. This means that the model uses the information from the previous batch to update its state and generate node embeddings for the current batch. This lag-one scheme helps maintain temporal consistency and ensures that the model captures the correct temporal patterns. Fig.~\ref{fig:train_dynamic} provides a graphical illustration of the training flow between epochs. The pseudo-code of the training procedure with cross-entropy is summarized in Algorithm~\ref{alg:train}.

\begin{algorithm}[H]
  \caption{Standard Training Procedure for TGNN}\label{algo:standard_train}
  \label{alg:train}
\begin{algorithmic}
   \STATE {\bfseries Initialization:} $H_{0} \leftarrow \Fb$ \COMMENT{Intialize embedding with feature vector of vertices and edge} \\
  \FOR{t=1 \textbf{to} T}
     \FOR{$B_i \in B_2,...,B_K$}
      \item   $B_i^{-} \leftarrow \text{Sample negative events}$ 
      \item   $\bar{B}_{i} = B_i^{-} \bigcup B_{i} $
      \item   $\bar{B}_{i-1} \leftarrow \text{Temporal batch from last iteration}$
      \item   $M_i = \mathrm{msg}(H_{i-1},\bar{B}_{i-1})$
      \item   $H_i = \mathrm{emb}(M_i,H_{i-1}, \mathcal{N}_i),$ \COMMENT{where $\mathcal{N}_i$ is the (Temporal) neighbourhood of vertex }
      \item   Compute the loss (e.g., binary cross-entropy) and run the training procedure (e.g., backpropagation) 
      $$\mathcal{L}(H_i, B_{i}) $$
     \ENDFOR
  \ENDFOR
\end{algorithmic}
\end{algorithm}    

\begin{algorithm}[!h]
  \caption{VCS}\label{algo:vcs}
  \label{alg:train2}
\begin{algorithmic}
   \STATE {\bfseries Input:}  $\Yb$, $\hat{\Yb}$\\
\STATE {\bfseries Initialization:}  $ \tau$
    \item Find the set of disagreement $\cE_{\mathrm{disg}}$ based on $\Yb$ and $\hat{\Yb}$
  \FOR{i=1 \textbf{to} $\tau$}
      \item  Compute $\mathrm{D}_{\mathrm{disg}}$ as 
    \begin{align*}
    \mathrm{D}_{\mathrm{disg}}^{(i)} &= \sum_{e \in \hat{\mathcal{E}}_{\mathrm{disg}}^{(i)}}\mathrm{d}(e,\mathcal{E}_{\mathrm{disg}}), \\
    \mathrm{d}(e,\mathcal{E}_{\mathrm{disg}}) &= \min \left\{|t_e-t_{e'}| \bigg| e' \in \mathcal{E}_{\mathrm{disg}}, e' \neq e \right\}.
    \end{align*}
      \item   generate a random set form $\cE_{\mathrm{r}}^{(i)}$ of the same cardinality as $\hat{\cE}_{\mathrm{disg}}^{(i)}$.
      \item  compute $\mathrm{D}_{\mathrm{r}}^{(i)}$ as above.
     \ENDFOR

   \item  Compute and return the VCS as
      \begin{align*}
    \mathrm{VCS} = | 1/2 -
     \mathcal{T}(\mathcal{E}_{\mathrm{disg}}, \mathcal{E}_{\mathrm{r}}, \tau)|,\\ 
     = \left | \frac{1}{2} - \frac{1}{\tau} \sum_{i  = 1}^{\tau} \frac{\mathrm{D}_{\mathrm{r}}^{(i)}}{\mathrm{D}_{\mathrm{r}}^{(i)}+\mathrm{D}_{\mathrm{disg}}^{(i)}} \right |.
\end{align*}
\end{algorithmic}
\end{algorithm}

\begin{algorithm}[!h]
  \caption{VCA}\label{algo:vca}
  \label{alg:train2}
\begin{algorithmic}
   \STATE {\bfseries Initialization:}  $H_{0} \leftarrow \Fb$ \COMMENT{Intialize embedding with feature vector of vertices and edge} \\
\STATE {\bfseries Initialization:}  $\gamma, \tau$
  \FOR{t=1 \textbf{to} T}
     \FOR{$B_i \in B_1,...,B_K$}
      \item   $B_i^{-} \leftarrow \text{Sample negative events}$ 
      \item   $\bar{B}_{i} = B_i^{-} \bigcup B_{i} $
      \item   $\bar{B}_{i-1} \leftarrow \text{Temporal batch from last iteration}$
       \item   $M_i = \mathrm{msg}(H_{i-1},\bar{B}_{i-1})$
      \item   $H_i = \mathrm{emb}(M_i,H_{i-1},A_i)$
      \item Get the prediction and ground truth for the batch $\hat{\Yb}(\bar{B}_{i-1})$, ${\Yb}(\bar{B}_{i-1})$
      \item Suppose the disagreement between prediction and ground truth is $k$, sample $k$ random samples from the dataset.
      \item   Compute the loss and run the training procedure (e.g., backpropagation) $$ \mathcal{L}(\bar{B}_{i-1}) + \gamma \cdot   \left \| \frac{1}{2} - \mathcal{T}_{\mathrm{soft}}(\mathcal{E}_{\mathrm{disg}}, \mathcal{E}_{\mathrm{r}},\tau) \right \|.$$
     \ENDFOR
  \ENDFOR
\end{algorithmic}
\end{algorithm}

\section{Additional Experiment Details}\label{appendix:exp}

\subsection{Hardware and Software}
All the experiments of this paper are conducted on the following machine

CPU: two Intel Xeon Gold 6230 2.1G, 20C/40T, 10.4GT/s, 27.5M Cache, Turbo, HT (125W) DDR4-2933

GPU: four NVIDIA Tesla V100 SXM2 32G GPU Accelerator for NV Link

Memory: 256GB (8 x 32GB) RDIMM, 3200MT/s, Dual Rank

OS: Ubuntu 18.04LTS

\subsection{Dataset}
\subsubsection{Description}
We use the following public datasets provided by the authors of JODIE~\cite{kumar2019jodie}. (1) Wikipedia dataset contains edits of Wikipedia pages by users. (2) Reddit dataset consists of users’ posts on subreddits. In these two datasets, edges are with 172-d feature vectors, and user nodes are with dynamic labels indicating if they get banned after some events. (3) MOOC dataset consists of actions done by students on online courses, and nodes with dynamic labels indicating if students drop out of courses. (4) LastFM dataset consists of events that users listen to songs. MOOC and LastFM datasets are non-attributed. The statistics of the datasets are summarized in Table~\ref{tab:data_description}.

\subsubsection{Liscence}
All the datasets used in this paper are from publicly available sources (public paper) without a license attached by the authors.

\begin{table*}[h!]
  \centering
  \caption{Detailed statistic of the datasets.
  }  
  {\large
   \setlength\tabcolsep{4pt}
    \begin{tabular}{c|ccccc}
    \toprule
        Datasets  &        Wikipedia  & Reddit & MOOC & LastFM    & GDELT      \\
            
    \midrule
           \# vertices & 9,227 & 10,984 & 7,144 & 1,980 & 16,682\\
           \# edges   & 157,474 & 672,447 & 411,749 & 1,293,103 & 1,912,909 \\
           \# edge features & 172 & 172 & 0 & 0 & 186 \\
    \bottomrule
    \end{tabular}%
   }
  \label{tab:data_description}%
\end{table*}%

\subsection{Model Description}
For our evaluation, we use six state-of-the-art TGNN models, with two models from each of the three categories of TGNNs mentioned: TGN~\citep{rossi2021tgn} \& Tiger~\citep{zhang2023tiger} (memory-based TGNNs), TCL~\citep{wang2021tcl} \& TGAT~\citep{xu2020tgat} (attention-based TGNNs), and JOIDE~\citep{kumar2019jodie} \& DyRep~\citep{trivedi2019dyrep} (RNN-based TGNNs). We adopt the implementation of these baselines from~\cite{zhou2022tgl,dgb_neurips_D&B_2022,huang2024temporal}.  
\begin{itemize}
\item DyRep~\citep{trivedi2019dyrep} is a temporal point process-based model that propagates interaction messages via Recurrent Neural Networks (RNNs) to update the node representations. It employs a temporal attention mechanism to model the weights of a given node’s neighbors.
\item JOIDE~\citep{kumar2019jodie} has an
update operation and a projection operation. The former utilizes two coupled RNNs to recursively
update the representation of the users and items. The latter predicts the future representation of a
node, while considering the elapsed time since its last interaction.
\item TGAT~\citep{xu2020tgat} aggregates features of temporal-topological neighborhood and temporal interactions
of dynamic network. The proposes TGAT layer employs a modified self-attention mechanism as
its building block where the positional encoding module is replaced by a functional time encoding.
\item TCL~\citep{wang2021tcl} employs a transformer module to generate temporal neighborhood representations for nodes involved in an interaction. It models the inter-dependencies with a co-attentional transformer at a semantic level. Specifically, TCL utilizes two separate encoders to extract representations from the temporal neighborhoods surrounding the two nodes involved in an edge.
\item TGN~\citep{rossi2021tgn} \& Tiger~\citep{zhang2023tiger} are memory-based TGNN framework for learning on continuous time dynamic graphs. It consists of the following components: memory module, message function, message aggregator, memory updater, and embedding module. Both TGN and Tiger updates the node memories at test time with newly observed edges. The main difference between these model reside in the employment of fresh restarted in Tiger.
\end{itemize}

\subsection{Implementation Details.}
We optimize all models using Adam and use supervised binary cross-entropy loss as the objective function. We train the models for 100 epochs and use an early stopping strategy with a patience of 10. We select the model that achieves the best performance on the validation set for testing. We set the learning rate and batch size to 0.001 and 200 for all methods on all datasets. We run the methods five times nd report the average performance to eliminate deviations. 

\end{document}